\documentclass[runningheads]{llncs}

\usepackage{eccv}

\usepackage{eccvabbrv}

\usepackage{graphicx}
\usepackage{booktabs}
\usepackage{multirow}
\usepackage[table]{xcolor}

\usepackage{algorithm}
\usepackage{algpseudocode}

\usepackage[accsupp]{axessibility}  
\usepackage{graphicx}

\usepackage{hyperref}

\usepackage{orcidlink}
\newcommand{\frameworkname}{\textsc{HAS}}

\begin{document}

\title{HAS: Highlight-guided Attention Steering for Multimodal LLM Video Summarization} 

\titlerunning{HAS: Attention Steering for Video Summarization}

\renewcommand{\thefootnote}{\fnsymbol{footnote}}
\setcounter{footnote}{0}
\author{Rui Chu\textsuperscript{*} \and
Yingjie Lao\textsuperscript{*}}

\authorrunning{R.\ Chu and Y.\ Lao}

\institute{Tufts University, Medford, MA, USA}

\maketitle
\footnotetext{Corresponding author, Email: rui.chu@tufts.edu}
\footnotetext{Corresponding author, Email: yingjie.lao@tufts.edu}
\setcounter{footnote}{0}
\renewcommand{\thefootnote}{\arabic{footnote}}

\begin{abstract}


Video understanding has become more and more important with the growth of Artificial Intelligence (AI) for video generation. Recently, Multimodal Large Language Model (M-LLM) has shown its capability in video understanding. Video summarization, a specific domain of video understanding, has proven its importance for efficient navigation and retrieval. Both video understanding and video summarization require a good selection of key frames in a video. 
Current video summarization methods heavily focus on the selected key frames and correlated segment captions. However, existing approaches overlook the perspective of treating the importance of the frames globally. 
We argue that using discrete selected frames for summarization will not only reduce the understanding coherence, but also lost important information in the video, as well as wasting the original capacity of the MLLMs.  In this paper, we propose \textbf{\frameworkname}, a \textbf{H}ighlight-guided \textbf{A}ttention \textbf{S}teering method for video summarization. We consider a challenging but practical setting where the video given to MLLMs for summarize should be continuous but with highlight guidance. \frameworkname~mainly consists of two parts:
The first part is to find a continuous frame-level highlight distribution for the video globally. 
The second part is to apply the highlight distribution as an attention steering vector for the MLLM, targeting a better understanding of the video, and thus during the model inference time, putting more attention on the highlighted frames, while avoiding lost entire information on less highlighted frames through putting less attention instead of forgetting them. 
We evaluated \frameworkname~on a variety of benchmarks, and it has shown convincing performance in video summarization.    
\end{abstract}

\section{Introduction}
\label{sec:intro}

Video understanding~\cite{buch2022revisiting} has become more and more important with the development of video generation~\cite{DBLP:journals/displays/ZhangSY26, DBLP:journals/ijcv/ZhaoDWYLCS26}, image and 3D generative visual modeling and robustness~\cite{DBLP:conf/cvpr/MengMSDL25, DBLP:conf/cvpr/HanZCL0L25}, knowledge retrieval and in context learning~\cite{chu2026debiasragtuningfreepathfair, chu2026bam}, animation rendering~\cite{DBLP:journals/corr/abs-2506-02733}, and interactive applications, since the video data has exceeded human capacity for consumption. Video summarization, a specific sub-domain of video understanding, requires an efficient processing method which \textit{firstly,} capture essential content (frames) and \textit{secondly,} process the content into concise summaries~\cite{DBLP:conf/cvpr/LeeGC25}.
Recent multimodal large language models (M-LLMs)~\cite{DBLP:journals/ieeejas/YuWWY26} process video as a sequence of visual tokens (time-ordered embeddings of frame-level tubelets) via cross-attention~\cite{DBLP:journals/access/LiaoLW26} and output instruction-following answers, making both video understanding and summarization feasible. 
Although MLLM can handle varieties of video understanding tasks, how to guide the model efficiently summarize the video during inference time while avoiding losingtoo much information is worth exploring.

Earlier video summarization through neural networks always require a training progress. Video summarization tasks, compared to traditional video understanding tasks, requires more consistency of the understanding of the entire video~\cite{DBLP:journals/ijcv/ChenHXPWCLLW26}. 
Many existing methods can summarize the main content, a common pipeline is still to first select discrete frames or segments (e.g., keyframes/keyshots) and then summarize based on the selected subset \cite{DBLP:conf/nips/NarasimhanRD21, DBLP:conf/cvpr/LeeGC25}. In brief, 1) finding the important frames; 2) understanding each frames, and summarize based on the combination of single frame understandings.
However, such hard selection overwhelmingly rely on discrete selected highlight frame sequence window~\cite{DBLP:conf/cvpr/LeeGC25}, which not only overlooked the MLLM capability of finding highlight through \textbf{attention} mechanism~\cite{DBLP:conf/nips/VaswaniSPUJGKP17}, but also impact the coherence of video understanding and summarization, harming downstream retrieval. 

Even though early works has made a video highlight into an import-score distribution coorelated to each frame time, using the highlight distribution to continuous enhance the video summarization has been neglected. 
In the Nature Language Processing (NLP) domain, inference-time attention steering has been studied as a way to guide a model to focus on user-specified important parts. For example, PASTA proposes a post-hoc attention steering approach that reweights attention at inference time, so the model can read emphasized tokens more like human readers.
With the development of multimodal LLMs (MLLMs), inference-time attention intervention has also been explored for vision-language settings. FarSight shows that modifying the decoding-time token interaction (via causal-mask-based intervention) can mitigate hallucinations in MLLMs, and it is effective on both image and video benchmarks \cite{Tang_2025_CVPR}. This suggests that inference-time manipulation of attention-related mechanisms can be a practical lever for improving multimodal reasoning and generation \cite{Tang_2025_CVPR}.

More broadly, lightweight inference-time control methods are popular, because they can adjust model behaviors without expensive finetuning. 
At the systems level, complementary advances in efficient~\cite{10.1145/3665314.3672281}, hardware-aware, and privacy-preserving AI computation also aim to make advanced model inference more practical under real deployment constraints \cite{10.1145/3676536.3697124, 10942749}.
A representative direction is vector/activation steering, where a steering vector is injected into hidden activations during generation to slightly shift the output distribution \cite{Konen_2024_EACL, Stolfo_2025_ICLR}. This is similar to human reading: if a reader has a guidance about what is important, it is easier to understand the content; in models, such steering signals can also act as a soft guidance to allocate computation and attention \cite{Konen_2024_EACL}.

Similarly, Considering the way of a real human summarizing a video is to \textit{firstly}, watch the entire video, and \textit{secondly}, recalling the video and summarizing by memorized more on important moments and less on less exciting moments~\cite{DBLP:journals/ais/TorrePB26}. The current work of video summarization~\cite{DBLP:conf/cvpr/LeeGC25} is more close the scenario to skimming the video first and summarize based on the skimmed frames, which can cause inconsistency. 
However, it is still unclear how to use a simple and controllable steering signal to guide a Video-MLLM to summarize videos more precisely, while keeping the video input continuous and avoiding losing video details. Meanwhile, the community has built query-conditioned highlight supervision for videos (e.g., clip-wise saliency scores conditioned on natural-language queries) \cite{Lei_2021_NeurIPS, Moon_2023_CVPR}, but these signals are mostly used for highlight detection or moment retrieval, not as an internal attention prior for generative video summarization. 
These observation motivate the research question we are trying to address: \textit{How to 1) have a highlight distribution of a video (acting as a teacher), and thus 2) using the highlight distribution to guide the M-LLM generate a more consistent and comprehensive video summarization?}

\begin{figure}
    \centering
    \includegraphics[width=1\linewidth]{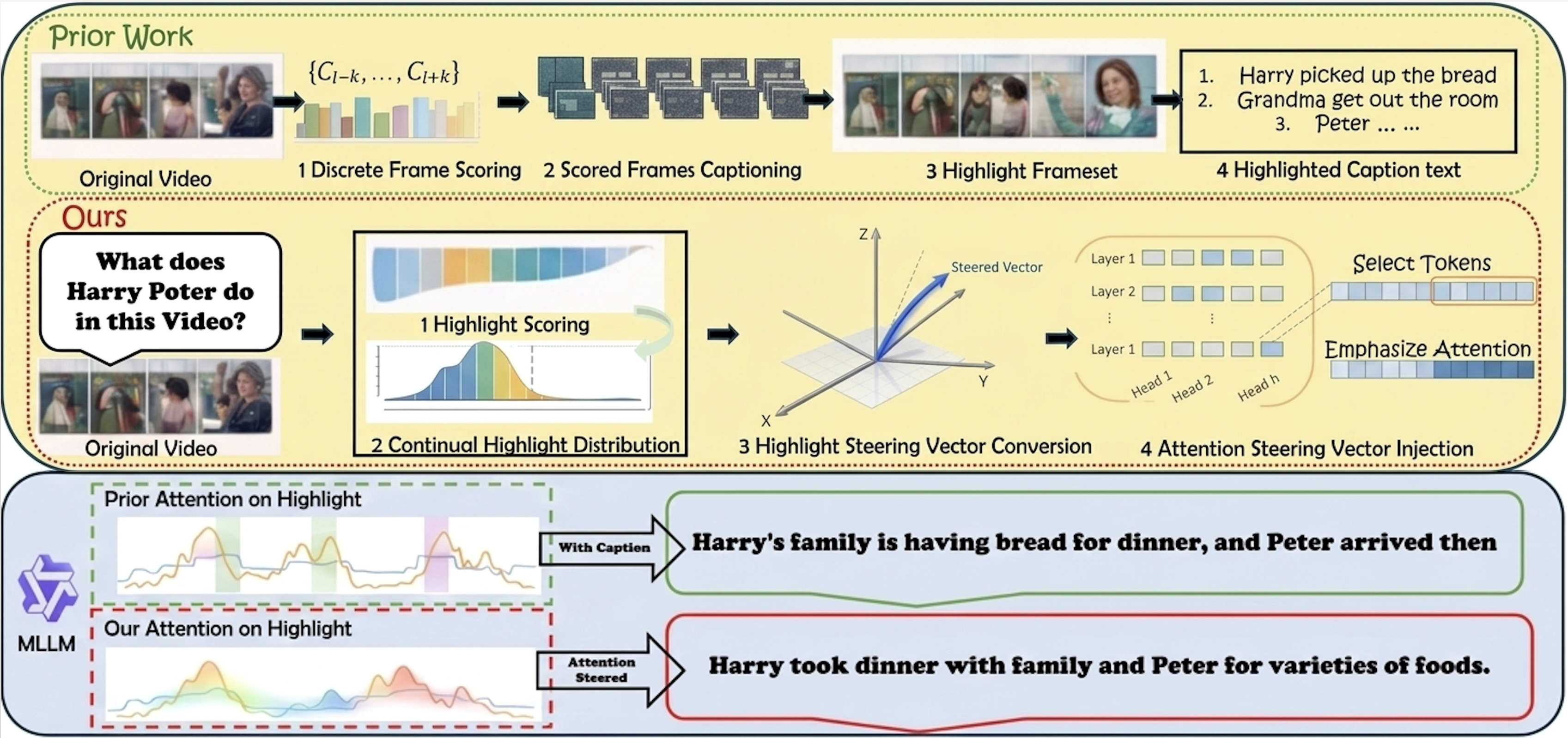}
    \caption{\textbf{Top:} Unlike prior works assign \emph{discrete} importance scores and \emph{hard-select} a few highlights before summarizing, \frameworkname preserves the full video context and treats highlighting as \emph{continual} attention guidance through steering vector. \textbf{Bottom:} While smoothly bias a frozen video MLLM toward highlight moments, \frameworkname~ does not neglect the peace time steps, better exploiting model capacity while reducing missed evidence for more coherent and faithful summaries.}
    \label{fig:placeholder}
\end{figure}

In this work, we propose \textbf{HAS}, a \emph{Highlight-guided Attention Steering} framework for video summarization. \frameworkname~inspired from both text summarization and image caption generation, where the performance can be improved through steering method. Meanwhile, ~\frameworkname is also inspired by the LLM quantization~\cite{DBLP:journals/access/ParkC26a} and distillation~\cite{DBLP:journals/eswa/HuangLCWLN26}, both aim to compress information while retaining essential knowledge: we consider highlighting video as a quantization progress, and use the highlight as a teacher signal to guide the MLLM generating progress, similar to a distillation process.  \frameworkname~consists of two parts: 1) introduces a smooth way to find a continuous highlight distribution for a selected video based on the entire frame sequence; 2) introduces a novel \textit{Attention Steering} method for MLLM to better use the highlight distribution as a steering vector to guide the attention of MLLM towards the high scoring pieces while remaining less attention towards other frames for final summarization. The high-level workflow can be illustrated in Fig.~\ref{fig:placeholder}.

Our contributions are summarized as followed,

\begin{itemize}
\item To the best of our knowledge, ~\frameworkname~ is the first work to use attention steering to guide the MLLM, and for a specific goal, video summarization.
\item \frameworkname~treat the video summarization from a continuous perspective, enhancing the consistency of LLM-based video summarization works.
\item \frameworkname~is a inference-time light-weight, backbone model training free framework, which is easy to be adapted towards pre-trained models.
\item Through extensive experiments, \frameworkname~ achieved an outstanding performance on both coherence, summarization performance and scalability.
\end{itemize}


\section{Related Works}
\label{sec:review}

    \subsection{Video summarization}
A standard formulation in video summarization is to first predict a frame- or clip-level importance curve and then convert it into hard keyframe or keyshot selection under a budget~\cite{DBLP:conf/cvpr/LeeGC25}. Early work in this line is mostly visual-only and relies on supervised temporal modeling; TVSum is a representative benchmark and formulation that summarizes videos from shot-level importance annotations \cite{song2015tvsum}. CLIP-It extends this paradigm to multimodal summarization, and unifies generic and query-focused settings by learning frame importance from a language-guided transformer \cite{DBLP:conf/nips/NarasimhanRD21}. Scaling Up Video Summarization Pretraining with Large Language Models further shows that LLM-generated supervision can improve summarization pretraining at scale \cite{argaw2024cvpr}. 
Recent efforts also expand the summarization setting beyond unimodal V2V: V2Xum-LLM unifies video-to-video, video-to-text, and joint cross-modal summarization via temporal prompt instruction tuning \cite{DBLP:conf/aaai/Hua0X025}. In parallel, VISTA builds a large-scale video-to-text benchmark for scientific presentations and highlights the challenges of long-form, domain-specific abstractive summarization.
More recent MLLM-based summarization still follows the same score-then-select pipeline: LLMVS converts frames into captions, estimates local importance with an LLM, and refines it with a global aggregator before summary construction \cite{DBLP:conf/cvpr/LeeGC25}. A related efficiency line also performs hard visual selection before downstream reasoning, including M-LLM Based Video Frame Selection, Flexible Frame Selection, Adaptive Keyframe Sampling, and the agentic AKeyS framework \cite{hu2025mllmfs, fan2025akeys}. A zero-shot alternative further shows that contrastive features can already approximate frame importance without task-specific retraining \cite{pang2024zeroshotvs}. Different from all these methods, we do not use the highlight signal as a final selector; we use a coarse continuous highlight distribution only as a steering prior for MLLM summarization.


  \subsection{Query-based highlight}
Query-based moment retrieval and highlight detection are closely related because they also learn a query-conditioned temporal relevance curve, but their output is still a moment boundary or a highlight score rather than a generated summary. 
Classic highlight detection methods (without language queries) also learn a temporal highlight score curve via ranking supervision, e.g., the pairwise deep ranking framework for first-person video summarization \cite{DBLP:conf/cvpr/YaoMR16}.
QVHighlights establishes this setting with natural-language queries, relevant moment annotations, and clip-level saliency labels \cite{Lei_2021_NeurIPS}. UMT unifies moment retrieval and highlight detection in a shared multimodal transformer. QD-DETR further learns query-dependent video representations through early cross-attention and negative video-query training, leading to better localization and saliency estimation \cite{Moon_2023_CVPR}. This line is the closest to our highlight prior, but the highlight curve is their final prediction target, while in our method it is only an intermediate signal that guides summarization.


  \subsection{Attention Steering}
    Another relevant line shows that attention can be modified at inference time as a practical control interface. Attention Biasing and Context Augmentation demonstrates that encoder-decoder transformers can be controlled in zero-shot fashion by directly biasing cross-attention during generation \cite{hazarika2022attentionbiasing}. PASTA makes this idea explicit for LLMs by reweighting selected attention heads so that the model attends more to user-emphasized text spans, without parameter updates. In multimodal and video generation, FarSight modifies the causal mask during decoding to improve visual token propagation and reduce hallucination \cite{Tang_2025_CVPR}. Closely related video-MLLM work also changes how visual tokens compete for attention: Vista-LLaMA adjusts relative token distance to prevent visual evidence from being ignored in long generations \cite{ma2024vistallama}, MASH-VLM disentangles spatial and temporal attention to reduce action-scene hallucination \cite{bae2025mashvlm}, and SEAL learns semantic attention over compact long-video units instead of dense raw frames. These works support the general paradigm that attention-level control is effective, but none of them injects a frame-level highlight distribution for video summarization.

    A final related question is how to construct and inject a reasonable highlighting prior. Highlight-Transformer provides direct evidence from text summarization: it introduces a highlighting matrix that explicitly increases attention weights on key phrases. Post-hoc attribution methods give another source of such priors. Excitation Backprop for RNNs localizes spatiotemporal evidence in video models without retraining \cite{bargal2018ebr}. Grad-CAM gives a simple gradient-based localization signal for visual backbones and is widely used as a saliency baseline \cite{selvaraju2017gradcam}. Extremal Perturbations estimates smooth masks by measuring which input regions most affect the prediction \cite{fong2019extremal}. These methods are mainly explanation tools rather than control mechanisms, but they support our assumption that a coarse continuous saliency signal can be extracted without dense labels and then reused to guide generation.

\section{Methodology}
\label{sec: method}

In this section, we present~\frameworkname~, the Highlight-guided Attention Steering method for video summarization. 

We begin by introducing the problem statement for video summarization in Section \ref{sec:problem statement} for better formulation, and then show the overview of our method in Section~\ref{sec:method_overview}, illustrating notations and explaining Figure.~\ref{fig:placeholder}. Afterwards, we illustrate \frameworkname~step by step in the following parts.


\subsection{Problem Statement}
\label{sec:problem statement}



The goal of video summarization is to find the best video summarization output text $\mathbf{O}$ through MLLM $\mathbf{M}$ given a selected video, considering as a sequence of frames $\mathbf{F}$;  and a user query/prompt $q$.
Let $\mathbf{F}=[\mathbf{F}_1,\ldots,\mathbf{F}_T]$ be a video sequence frame, $T$ denotes the temporal length of the video. 
We are aiming at using continuous video information instead of discrete frame pieces for outputing best $\mathbf{O}$, as prior LLM-based video summarization works did~\cite{DBLP:conf/cvpr/LeeGC25}, and maintaining best original capacity of MLLMs. 

To this end, the optimization goal is to find the proper \textit{attention steering} vector $\mathbf{V}$ for $\mathbf{M}$ so that during inference time, best guide MLLM $\mathbf{M_V}$ through steering to generate the best optimized summarization output text $\mathbf{O}^*$, where
\begin{equation}
\label{eq:video_summarization_objective}
\mathbf{O}^* = \mathbf{M_V}(\mathbf{F},q)
\end{equation}

\subsection{Method Overview}
\label{sec:method_overview}

As shown in Fig.~\ref{fig:placeholder}, \frameworkname~consists of two stages:
(i) constructing a prompt-conditioned temporal continuous highlight distribution prior from the input video,
and (ii) converting the distribution into vectors and injecting into a MLLM at inference time via a attention steering policy.


\paragraph{Highlight construction }
Given a video $\mathbf{F}$ and $q$,
we first obtain a \emph{raw} highlight score sequence
$\hat{\mathbf{h}}=[\hat{h}_1,\ldots,\hat{h}_T]$ using an off-the-shelf highlight/moment module.
We then \emph{calibrate} it to a smooth and bounded highlight distribution
$\mathbf{h}=[h_1,\ldots,h_T]\in[0,1]^T$ 
by applying temporal smoothing.

\paragraph{Attention Steering Injection}
We map the distribution to visual tokens and form a token-level attention 
(denoted as $\mathbf{V}$) for the Video-MLLM.
The injection is governed by a small set of steering parameters following,
which we tune/choose on a validation set for stable and effective summarization.

\subsection{Highlight Distribution}
\label{sec:highlight_distribution}

\paragraph{Highlight generator.}
We reuse an off-the-shelf highlight generator $\mathcal{H}$ and only calibrate its output for steering. Given the input video $\mathbf{F}$ and prompt $q$, the generator produces a raw temporal score sequence $\hat{\mathbf{h}}=\mathcal{H}_{}(\mathbf{F}, q)$. When the generator outputs clip-level scores, we linearly interpolate them to length $T$. 

\paragraph{Temporal calibration.}
Raw highlight curves are noisy and not continuous (smooth) because it is based on the frame selection and length of frame time window. 
In \frameworkname~ needs to be continuous enough to guide attention. Thus, we firstly align the time frame the required $T$; and then, we normalize the score into the range of $[0,1]$ (line 1 to 3 in Algorithm.~\ref{alg:highlight_calibration}). 
We solve a lightweight one-dimensional calibration problem on top of the off-the-shelf scores.

The final result is a bounded and continuous highlight distribution $\mathbf{h}=[h_1,\ldots,h_T]$, which will be converted into steering vectors in Section~\ref{sec:steering_vector}.

\begin{algorithm}[t]
\caption{Prompt-conditioned highlight steering vector generation}
\label{alg:highlight_calibration}
\begin{algorithmic}[1]
\Require video frames $\mathbf{F}=[F_1,\ldots,F_T]$,  Highlight Identifying Module $\mathcal{H}$,  query $q$.
\State $\hat{\mathbf{h}} \gets \mathcal{H}(\mathbf{F},q)$
    \State $\hat{\mathbf{h}} \leftarrow \mathrm{Interp}(\hat{\mathbf{h}}, T)$  
\State ${\mathbf{h}} \leftarrow \mathrm{MinMaxNorm}(\hat{\mathbf{h}})$
\State $\mathbf{V} \leftarrow  \mathrm{Vectorize} (\mathbf{h})$ 
\State \Return $\mathbf{V}$
\end{algorithmic}
\end{algorithm}

\subsection{From Highlight Distribution to Steering Vector}
\label{sec:steering_vector}


As it is shown in line 4 of Algorithm.~\ref{alg:highlight_calibration}, vectorization is needed.
Since the target MLLM attends over visual tokens rather than scalar frame scores, \frameworkname~ lifts this frame-level distribution to the token level before attention intervention. Following prior work that converts explicit highlighting or external emphasis into attention bias or attention reweighting \cite{hazarika2022attentionbiasing}, we repeat each frame score over the $P$ visual tokens extracted from that frame and form the steering vector
\begin{equation}
\label{eq:steering_vector}
\mathbf{V}=\log\!\left(\mathrm{Repeat}(\mathbf{h},P)+\epsilon\right)\in\mathbb{R}^{TP},
\end{equation}
where $\mathrm{Repeat}(\mathbf{h},P)$ copies $h_t$ to all $P$ visual tokens of frame $F_t$. 
We add a small constant $\epsilon$ for numerical stability, which avoids collapsing low-highlight frames to $-\infty$ in log-space and keeps HAS as soft steering rather than hard frame selection. 
The logarithm makes $\mathbf{V}$ directly usable as an additive bias on attention logits in the next subsection. 
(Similar inference-time attention intervention has also been shown effective in MLLMs \cite{Tang_2025_CVPR}).

\subsection{Inference-time Attention Steering}
\label{sec:attention_steering}

\begin{algorithm}[t]
\caption{HAS inference-time attention steering}
\label{alg:has_steering}
\begin{algorithmic}[1]
\Require video frames $\mathbf{F}$, prompt $q$, frozen MLLM $M$, Steering Vector $\mathbf{V}$, Attention steering head set $\mathcal{S}$, gate parameters $\{a_{\ell,m}\}$, strengths $\{\beta_{\ell,m}\}$
\State Encode $\mathbf{F}$ into visual tokens and feed them to $M$
\For{each decoding step}
    \For{each selected cross-attention head $(\ell,m)\in\mathcal{S}$}
        \State Compute pre-softmax attention logits $\mathbf{A}^{(\ell,m)}$
        \State $g_{\ell,m}\leftarrow \sigma(a_{\ell,m})$
        \For{each query row $i$}
            \State $\mathbf{A}^{(\ell,m)}_{i,:} \leftarrow \mathbf{A}^{(\ell,m)}_{i,:} + g_{\ell,m}\beta_{\ell,m}\mathbf{V}$
        \EndFor
        \State Compute steered attention with $\mathrm{softmax}(\mathbf{A}^{(\ell,m)})$
    \EndFor
    \State Decode the next token
\EndFor
\State \Return summary $\mathbf{O}^*$
\end{algorithmic}
\end{algorithm}

Given the steering vector $\mathbf{V}$, \frameworkname intervenes on a small subset of visual cross-attention heads during decoding. This follows the same inference-time steering spirit as attention biasing~\cite{hazarika2022attentionbiasing} and PASTA, and is also consistent with the plug-and-play attention intervention view of FarSight~\cite{Tang_2025_CVPR}. 
For a selected head $(\ell,m)\in\mathcal{S}$, let $\mathbf{A}^{(\ell,m)}$ denote the pre-softmax attention logits from the current text queries to all visual tokens. To enable \emph{differentiable} policy learning over heads, we introduce a continuous gate $g_{\ell,m}=\sigma(a_{\ell,m})\in[0,1]$ for each candidate head. \frameworkname applies a row-wise additive bias
\begin{equation}
\label{eq:has_logit_bias}
\tilde{\mathbf{A}}^{(\ell,m)}_{i,:}=\mathbf{A}^{(\ell,m)}_{i,:}+g_{\ell,m}\,\beta_{\ell,m}\mathbf{V}, \qquad (\ell,m)\in\mathcal{S},
\end{equation}
and leaves all other heads unchanged, where $\ell$ and $m$ index the transformer layer and attention head, $\mathcal{S}$ denotes the candidate head set, $\sigma(\cdot)$ is the sigmoid function, $\beta_{\ell,m}$ is the head-wise steering strength, and $i$ indexes the query row (with $:\,$ spanning all visual tokens).
Since $\mathbf{V}$ is constructed in log-space, Eq.~\eqref{eq:has_logit_bias} is equivalent to multiplicative reweighting of the unnormalized attention scores, but is easier to stabilize and easier to combine with existing causal or padding masks. The small constant $\epsilon$ in Eq.~\eqref{eq:steering_vector} avoids collapsing low-highlight frames to $-\infty$ in log-space, so \frameworkname remains a soft steering method rather than hard frame selection.
Alg.~\ref{alg:has_steering} implements this intervention inside the standard autoregressive decoding loop: after encoding $\mathbf{F}$ into visual tokens (line~1), each generation step computes $\mathbf{A}^{(\ell,m)}$ (line~4), applies the gated bias in Eq.~\eqref{eq:has_logit_bias} for heads in $\mathcal{S}$ (line~5--8), and then proceeds with the usual softmax and token decoding using the modified logits (line~9--13).

The optimization target of this stage is only the steering policy $\Theta=\{a,\beta\}$, not the MLLM nor the highlight generator. We choose $\Theta$ on a held-out calibration/validation set by minimizing the standard teacher-forcing negative log-likelihood of the reference summaries~\cite{DBLP:conf/nips/SutskeverVL14, DBLP:conf/nips/VaswaniSPUJGKP17}, while keeping the whole MLLM frozen, and we add a sparsity term to encourage using only a small subset of heads, following the common practice of sparse head selection for interpretability and regularization:
\begin{equation}
\label{eq:has_policy_loss}
\min_{a,\beta}\ \mathcal{L}_{\mathrm{NLL}}(a,\beta)\;+\;\lambda_s\sum_{(\ell,m)\in\mathcal{S}} g_{\ell,m}.
\end{equation}
Here $\lambda_s$ is the sparsity weight that trades off summary likelihood and the number of activated heads.
In practice, we optimize $\{a_{\ell,m},\beta_{\ell,m}\}$ by minimizing the teacher-forcing NLL in Eq.~\eqref{eq:has_policy_loss} with a standard first-order optimizer (Adam~\cite{DBLP:journals/corr/KingmaB14}), while keeping the whole MLLM frozen. 
We optimize $\{a,\beta\}$ once on a held-out set; at test time, $\{a,\beta\}$ are fixed and thus Alg.~\ref{alg:has_steering} will performs forward-only steering.

\section{Experiments}
\label{sec:experiments}


In this section, we evaluate ~\frameworkname~ across a range of video summarization tasks. 
All experiments are conducted on Nvidia L40S GPUs.

\subsection{Experimental Settings}
\label{sec:exp_settings}

\subsubsection{Datasets and benchmarks.}
We evaluate ~\frameworkname~ on three benchmark families that align with our cross-comparison tables.
\textbf{(i) V2V summarization} on \textsc{SumMe}~\cite{gygli2014summe} and \textsc{TVSum}~\cite{song2015tvsum} measures the quality of frame/shot importance ranking against human annotations.
\textbf{(ii) Cross-modal summarization} on \textsc{VideoXum}~\cite{lin2024videoxum} evaluates \emph{video-to-text (V2T)}, \emph{video-to-video (V2V)}, and \emph{joint video\,+\,text (V2VT)} summaries in a unified protocol.
\textbf{(iii) Scientific video-to-text summarization} on \textsc{VISTA}~\cite{DBLP:conf/acl/0001WYMSZQLD25} benchmarks long-form academic talk summarization, emphasizing both semantic quality and factual grounding.

\subsubsection{Models.}
To avoid conclusions tied to a single backbone, we evaluate \frameworkname~ as a \emph{plug-and-play} inference-time module on a diverse set of \emph{open-source} video MLLMs.
Specifically, our backbone $\mathcal{M}$ is instantiated with:
(i) \textbf{mPLUG-Owl3}~\cite{mplugowl3},
(ii) \textbf{LLaVA-NeXT-Interleave}~\cite{llava-next},
(iii) \textbf{Video-LLaVA},
(iv) \textbf{LLaMA-VID},
(v) \textbf{Video-ChatGPT}~\cite{video-chatgpt},
and (vi) \textbf{Video-LLaMA}.
All backbones use publicly available implementations and checkpoints released by the authors (subject to the corresponding base-model licenses).
Unless stated otherwise, we keep $\mathbf{M}$ \emph{frozen} and apply \frameworkname~ purely at inference time.


\subsubsection{Baselines.}
For \textbf{V2V summarization}, we compare against:
\textbf{Visual-only} summarizers including VASNet~\cite{fajtl2019vasnet}, DSNet (anchor-based/anchor-free)~\cite{zhu2021dsnet}, DMASum, PGL-SUM~\cite{apostolidis2021pglsum}, MSVA~\cite{ghauri2021msva}, iPTNet, and CSTA~\cite{son2024csta};
\textbf{Visual+Text} methods CLIP-It~\cite{DBLP:conf/nips/NarasimhanRD21}, A2Summ~\cite{he2023a2summ}, SSPVS, and the large-scale pretraining baseline by Argaw et al.~\cite{argaw2024cvpr};
and \textbf{LLM-centric} methods including the zero-shot LLM scoring baseline and LLMVS~\cite{DBLP:conf/cvpr/LeeGC25}, as well as V2Xum-LLaMA from V2Xum-LLM~\cite{DBLP:conf/aaai/Hua0X025}.
For \textbf{cross-modal summarization} on \textsc{VideoXum}, we follow the established baselines and protocols from VideoXum~\cite{lin2024videoxum} and V2Xum-LLM~\cite{DBLP:conf/aaai/Hua0X025} (e.g., BLIP/Vid2Seq-based variants and V2Xum-LLaMA).
For \textbf{scientific summarization}, we report representative model families and settings (zero-shot and fine-tuning) as benchmarked by VISTA~\cite{DBLP:conf/acl/0001WYMSZQLD25}.

\subsubsection{Metrics.}
We report official metrics for each benchmark family.

\textbf{SumMe/TVSum}, we use Kendall's $\tau$ and Spearman's $\rho$ rank correlations (higher is better), following prior practice and LLMVS~\cite{DBLP:conf/cvpr/LeeGC25}.
For \textbf{VideoXum}, we evaluate V2T by BLEU-4 / METEOR / ROUGE-L / CIDEr, V2V by F1 / Spearman / Kendall, and V2VT by semantic alignment metrics (FCLIP / Cross-FCLIP)~\cite{lin2024videoxum, DBLP:conf/aaai/Hua0X025}.
For \textbf{VISTA}, we report ROUGE-Lsum (RLsum), BERTScore, and two video-grounded quality metrics VideoScore and FactVC~\cite{DBLP:conf/acl/0001WYMSZQLD25}.
Unless noted otherwise, larger values indicate better performance.

\subsection{Main Performance}

\subsubsection{Human-Aligned Temporal Importance Estimation}

\begin{table}[t]
\centering
\small
\setlength{\tabcolsep}{4pt}
\caption{\textbf{Human-aligned temporal importance estimation.}
Rank correlation ($\tau/\rho$; higher is better) between predicted importance trajectories and human annotations on V2V summarization.
}
\label{tab:v2v_summe_tvsum}
\begin{tabular}{lcccc}
\toprule
& \multicolumn{2}{c}{SumMe} & \multicolumn{2}{c}{TVSum} \\
Method & $\tau$ & $\rho$ & $\tau$ & $\rho$ \\
\midrule
Random & 0.000 & 0.000 & 0.000 & 0.000 \\
\midrule
\multicolumn{5}{l}{\textit{Visual}} \\
VASNet & 0.160 & 0.170 & 0.160 & 0.170 \\
DSNet-AB & 0.051 & 0.059 & 0.108 & 0.129 \\
DSNet-AF & 0.037 & 0.046 & 0.113 & 0.138 \\
DMASum & 0.063 & 0.089 & 0.203 & 0.267 \\
PGL-SUM & -- & -- & 0.206 & 0.157 \\
MSVA & 0.200 & 0.230 & 0.190 & 0.210 \\
iPTNet & 0.101 & 0.119 & 0.134 & 0.163 \\
CSTA & 0.246 & 0.274 & 0.194 & 0.255 \\
\midrule
\multicolumn{5}{l}{\textit{Visual + Text}} \\
CLIP-It & -- & -- & 0.108 & 0.147 \\
A2Summ & 0.108 & 0.129 & 0.137 & 0.165 \\
SSPVS & 0.192 & 0.257 & 0.181 & 0.238 \\
Argaw et al. & 0.130 & 0.152 & 0.155 & 0.186 \\
\midrule
\multicolumn{5}{l}{\textit{LLM-centric}} \\
LLM (zero-shot) & 0.170 & 0.189 & 0.051 & 0.056 \\
LLMVS & 0.253 & 0.282 & 0.211 & 0.275 \\
V2Xum-LLaMA & 0.296 & \textbf{0.378} & 0.222 & 0.293 \\
\midrule
\frameworkname~(ours) & \shortstack{\textbf{0.298}\\[-2pt]{\scriptsize $\pm$0.02}} & \shortstack{0.335\\[-2pt]
{\scriptsize $\pm$0.03}}& \shortstack{\textbf{0.224}\\[-2pt]{\scriptsize $\pm$0.02}} & \shortstack{\textbf{0.299}\\[-2pt]
{\scriptsize $\pm$0.03}}\\
\bottomrule
\end{tabular}
\end{table}

\paragraph{Human-aligned temporal importance estimation.}
Table~\ref{tab:v2v_summe_tvsum} shows that \frameworkname better aligns temporal importance with human annotations under both Kendall's $\tau$ and Spearman's $\rho$.
The gains over score-then-select pipelines suggest that inference-time steering with a calibrated temporal prior improves \emph{global} salience ordering rather than applying a trivial re-scaling.
Overall, \frameworkname narrows the gap between classical visual summarizers and LLM-centric approaches while remaining training-free.

\subsubsection{Cross-Modal Consistency without Sacrificing Text Quality}

\begin{table*}[t]
\centering
\scriptsize
\setlength{\tabcolsep}{3pt}
\caption{\textbf{Joint cross-modal summarization and video--text consistency.}
We evaluate V2T text quality (BLEU-4/METEOR/ROUGE-L/CIDEr), V2V temporal salience (F1/Spearman/Kendall), and V2VT semantic alignment (FCLIP/Cross-FCLIP); higher is better.
\textbf{}}
\label{tab:videoxum}
\begin{tabular}{lcccccccccc}
\toprule
& \multicolumn{4}{c}{V2T} & \multicolumn{3}{c}{V2V} & \multicolumn{2}{c}{V2VT} \\
Method & B-4 & M & R-L & C & F1 & Spr. & Kend. & FCLIP & Cross-FCLIP \\
\midrule
Frozen-BLIP & 0.0 & 0.4 & 1.4 & 0.0 & 16.1 & 0.011 & 0.008 & -- & -- \\
Vid2Seq-HCY & 2.3 & 8.2 & 19.0 & 7.6 & 24.2 & -- & -- & 0.888 & 0.214 \\
Vid2Seq-HC & 2.7 & 8.5 & 19.8 & 8.4 & 24.5 & -- & -- & 0.892 & 0.217 \\
Vid2Seq-HCV & 2.7 & 8.4 & 19.8 & 8.3 & 25.1 & -- & -- & 0.899 & 0.200 \\
VSUM-BLIP & -- & -- & -- & -- & 21.7 & 0.207 & 0.131 & -- & -- \\
TSUM-BLIP & 5.6 & 11.8 & 24.9 & 20.9 & -- & -- & -- & -- & -- \\
VTSUM-BLIP & 5.8 & 12.2 & 25.1 & 23.1 & 23.5 & 0.258 & 0.196 & 0.894 & 0.247 \\
V2Xum-LLaMA-7B & 5.8 & 12.3 & 26.3 & 26.9 & 29.0 & 0.298 & 0.204 & 0.931 & 0.253 \\
V2Xum-LLaMA-13B & 5.7 & 12.3 & 26.2 & 25.3 & 31.6 & 0.276 & 0.200 & 0.957 & 0.251 \\
\midrule
\frameworkname\ (ours) & \shortstack{\textbf{6.0}\\[-2pt]{\scriptsize $\pm$0.03}} & \shortstack{12.7\\[-2pt]{\scriptsize $\pm$0.3}} & \shortstack{\textbf{26.7}\\[-2pt]{\scriptsize $\pm$0.3}} & \shortstack{\textbf{28.0}\\[-2pt]{\scriptsize $\pm$0.8}} & \shortstack{32.0\\[-2pt]{\scriptsize $\pm$0.5}} & \shortstack{0.304\\[-2pt]{\scriptsize $\pm$0.02}} & \shortstack{0.230\\[-2pt]{\scriptsize $\pm$0.02}} & \shortstack{\textbf{0.963}\\[-2pt]{\scriptsize $\pm$0.01}} & \shortstack{\textbf{0.258}\\[-2pt]{\scriptsize $\pm$0.007}} \\
\bottomrule
\end{tabular}
\end{table*}

\paragraph{Cross-modal summarization.}
Table~\ref{tab:videoxum} indicates that \frameworkname improves cross-modal summarization in a balanced way: text quality (V2T) is maintained while temporal salience (V2V) and video--text consistency (V2VT) are strengthened.

The gain on semantic alignment metrics (FCLIP/Cross-FCLIP) suggests that \frameworkname better ties generated summaries to evidence-bearing segments even when exact temporal boundaries are ambiguous.This supports soft steering as a lightweight alternative to discrete extraction for joint V2T/V2V/V2VT evaluation.

\subsubsection{Faithful Summarization under Long-Context Evidence}

\begin{table}[t]
\centering
\small
\setlength{\tabcolsep}{4pt}
\caption{\textbf{Grounded long-form summarization.}
We report summary quality (RLsum/BERTScore) together with video-grounded alignment (VideoScore) and factual consistency (FactVC); higher is better.
}
\label{tab:vista_compact}
\begin{tabular}{lcccc}
\toprule
Method (setting) & RLsum & BERTScore & VideoScore & FactVC \\
\midrule
GPT-o1 (zero-shot) & 24.37 & 82.63 & 2.17 & 51.36 \\
Gemini 2.0 (zero-shot) & 24.29 & 82.64 & 2.02 & 52.02 \\
LLaVA-NeXT-Interleave (zero-shot) & 22.68 & 81.40 & 1.73 & 40.12 \\
mPLUG-Owl3 (zero-shot) & 22.84 & 81.39 & 1.77 & 42.07 \\
Plan-mPLUG-Owl3$^\star$ (zero-shot) & 22.97 & 81.45 & 1.86 & 47.37 \\
\midrule
mPLUG-Owl3 (full FineTune) & 32.91 & 84.22 & 3.28 & 71.94 \\
Plan-mPLUG-Owl3 (full FineTune) & \textbf{33.25} & \textbf{84.37} & \textbf{3.33} & \textbf{75.41} \\
\midrule
\frameworkname\ (ours, averaged on selected backbones) & 23.94 & 82.05 & 2.03 & 50.11 \\
\bottomrule
\end{tabular}
\end{table}

\paragraph{Grounded long-form summarization.}
Table~\ref{tab:vista_compact} shows that \frameworkname improves groundedness on long scientific talks: video--text alignment and factual consistency trend upward without sacrificing overall summary quality.
This suggests that calibrated temporal steering encourages evidence usage during generation, reducing unsupported details that commonly arise in long-context settings.
Taken together, \frameworkname offers a practical inference-time mechanism for faithful summarization in high-stakes domains.

\subsection{Coverage Performance}



\begin{figure*}[t]
  \centering
  \begin{subfigure}[t]{0.49\textwidth}
    \centering
    \includegraphics[width=\linewidth]{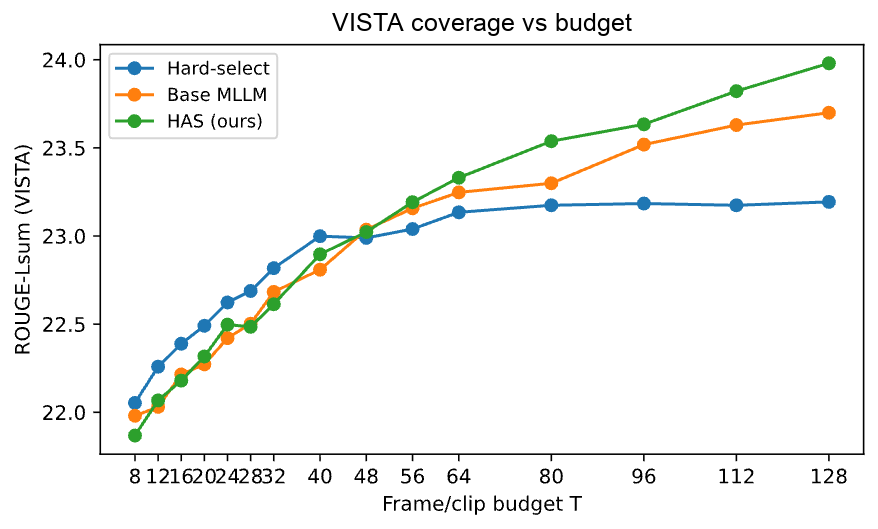}      
    \caption{Coverage (ROUGE-Lsum) vs. budget $T$.}
    \label{fig:vista_rlsum}
  \end{subfigure}\hfill
  \begin{subfigure}[t]{0.49\textwidth}
    \centering
    \includegraphics[width=\linewidth]{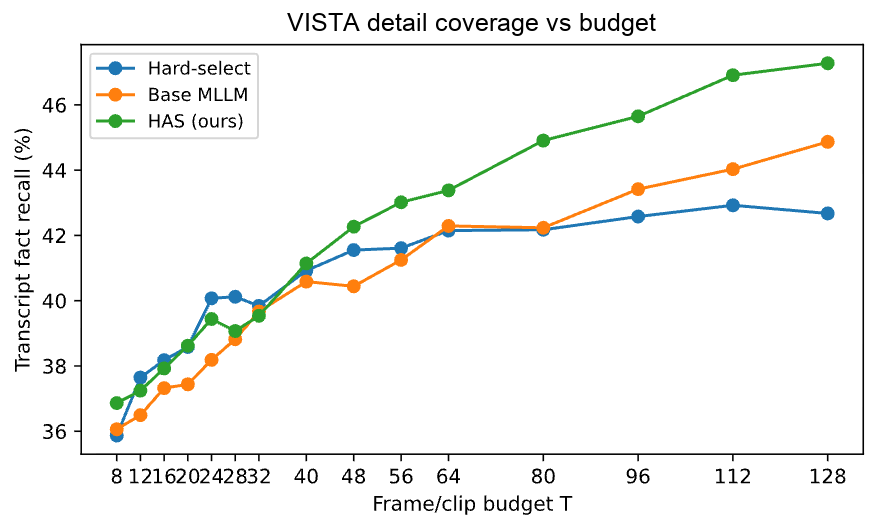}
    \caption{Coverage (fact-unit recall) vs. budget $T$.}
    \label{fig:vista_factrecall}
  \end{subfigure}

  \caption{\textbf{Coverage vs. budget on VISTA.}
(a) ROUGE-Lsum and (b) transcript fact-unit recall versus $T$.
Hard selection saturates early, while \frameworkname keeps improving at mid-to-high budgets, indicating reduced information loss under fixed-budget inference.}
  \label{fig:vista_coverage_ab}
\end{figure*}



\subsubsection {Coverage and information retention on VISTA.}

Fig.~\ref{fig:vista_coverage_ab} exhibits a shared saturation trend: marginal gains diminish as $T$ increases.
However, the discrete extraction baseline plateaus substantially earlier, revealing an irreversible bottleneck---once low-scored segments are removed, additional budget cannot restore missing evidence.
\frameworkname shows delayed saturation and a clear divergence from hard selection at mid-to-high budgets, suggesting improved retention of dispersed details common in long scientific talks.
The consistent separation on both ROUGE-Lsum and transcript fact recall further indicates that the gains reflect improved coverage rather than metric-specific tuning.
Together, these trends support our hypothesis that soft attention steering reduces information loss under fixed-budget inference.


\subsection{Scalability}





\begin{figure}[t]
  \centering

  \begin{minipage}[t]{0.48\linewidth}
    \vspace{0pt} 
    \centering
    \small
    \setlength{\tabcolsep}{5pt}

    \captionof{table}{\textbf{Zero-shot evaluation on MR.HiSum.}
    Following LLMVS~\cite{DBLP:conf/cvpr/LeeGC25}, models are trained on SumMe and directly evaluated on a chosen subset of 50 MR.HiSum videos.
    Higher is better.}
    \label{tab:mrhisum_zeroshot}

    \begin{tabular}{lcc}
      \toprule
      Method & $\tau$ & $\rho$ \\
      \midrule
      VASNet~\cite{fajtl2019vasnet} & 0.364 & 0.364 \\
      PGL-SUM~\cite{apostolidis2021pglsum} & 0.375 & 0.375 \\
      DSNet-AB~\cite{zhu2021dsnet} & 0.362 & 0.362 \\
      DSNet-AF~\cite{zhu2021dsnet} & 0.342 & 0.342 \\
      LLMVS~\cite{DBLP:conf/cvpr/LeeGC25} & 0.440 & 0.440 \\
      \midrule
      \textbf{HAS (ours)} & \textbf{0.45} & \textbf{0.45} \\
      \bottomrule
    \end{tabular}
  \end{minipage}
  \hfill
  \begin{minipage}[t]{0.48\linewidth}
    \vspace{0pt} 
    \centering
    \includegraphics[width=\linewidth]{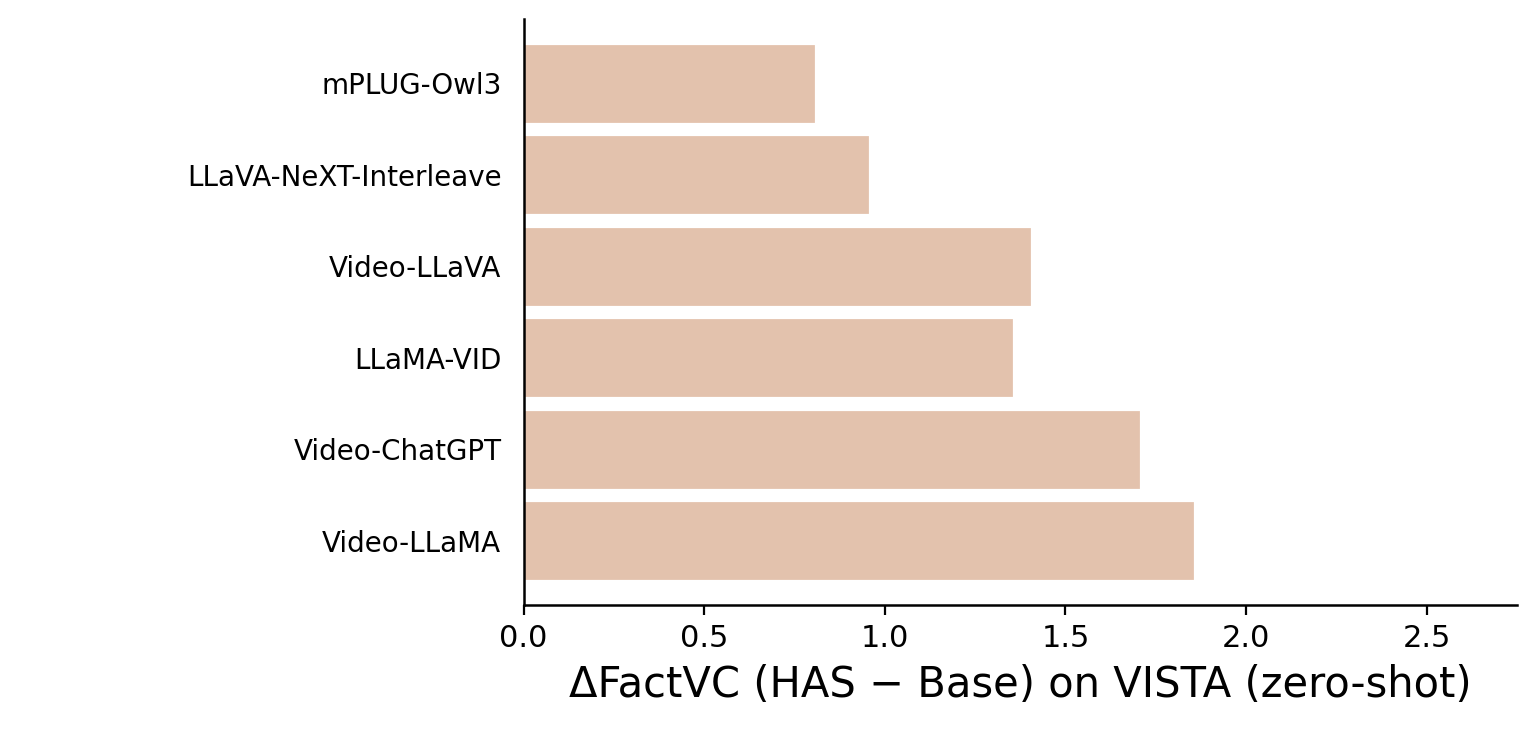}
    \caption{\textbf{Gains across open-source MLLMs.}
The improvement (Averaged) in groundedness on VISTA after adding \frameworkname\ to all selected open-source video MLLM backbone.
Positive values indicate that \frameworkname\ increases factual consistency.}
\label{fig:opensource_mmllm_delta}
  \end{minipage}


\end{figure}

\subsubsection{Zero-shot generalization}
Table~\ref{tab:mrhisum_zeroshot} reports a zero-shot transfer from SumMe to MR.HiSum without adaptation.
\frameworkname remains competitive under this distribution shift and improves both Kendall's $\tau$ and Spearman's $\rho$, suggesting more stable global importance ordering rather than overfitting to training-set idiosyncrasies.
These results indicate that calibrated temporal attention steering is robust and practical when retraining on the target domain is infeasible.

\subsubsection{Performance Across Different MLLMs.}
Fig.~\ref{fig:opensource_mmllm_delta} shows that \frameworkname\ yields consistent positive $\Delta$FactVC across a diverse set of open-source video MLLMs under the same zero-shot protocol.
This directly addresses the single-backbone concern: the gains are not tied to a particular architecture or training recipe, but stem from the inference-time steering mechanism.
Moreover, the variation in $\Delta$ across backbones suggests that \frameworkname\ acts as a complementary control layer---often providing larger benefits when the base model is more prone to missing or misusing evidence---while remaining beneficial even for stronger interleaving-style backbones.
Overall, these results support \frameworkname\ as a plug-and-play module that improves grounded long-form summarization in a backbone-agnostic manner.

\subsection{Controllable Highlight Distribution Visualization given Queries}

\begin{figure}[t]
    \centering
    \includegraphics[width=1\linewidth]{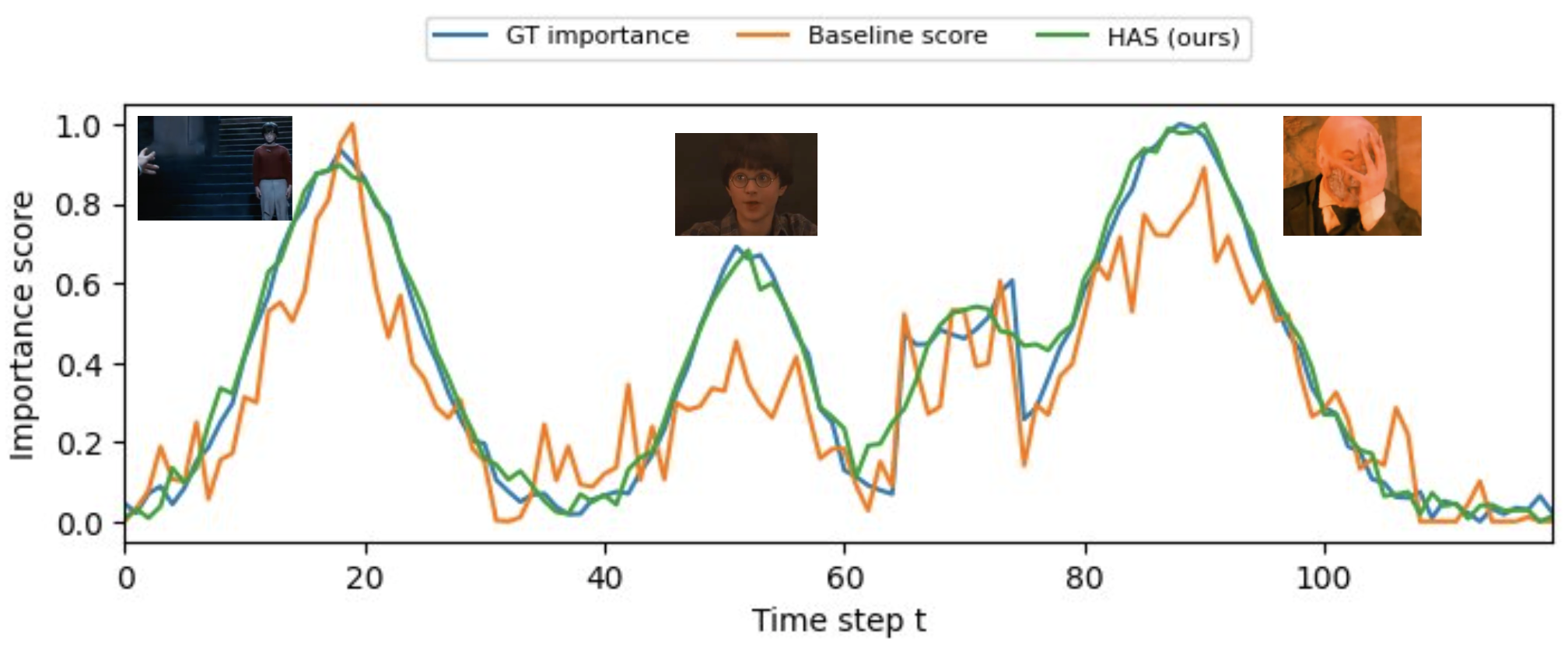}\\[2pt]
    {\small\textbf{(A)} Qualitative importance curves}\\[6pt]
    \includegraphics[width=1\linewidth]{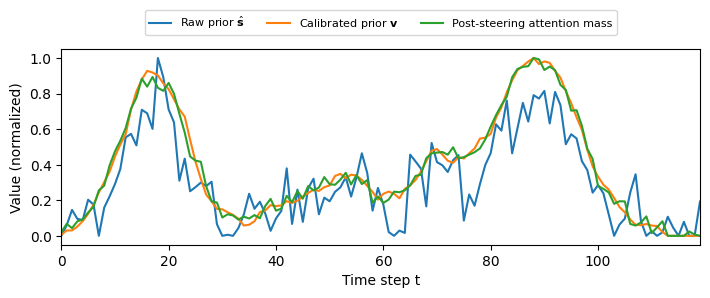}\\[2pt]
    {\small\textbf{(B)} Mechanistic visualization of \frameworkname~}
    \caption{Qualitative and mechanistic visualizations of \frameworkname~}
    \label{fig:qual_curves_ab}
\end{figure}

\subsubsection{Qualitative importance curves.}
Fig.~\ref{fig:qual_curves_ab}(A) compares the predicted importance trajectory with ground-truth (GT) scores.
Compared to the baseline, \frameworkname aligns salient peaks more accurately in both timing and duration while suppressing spurious oscillations on less important segments, consistent with the higher Kendall's $\tau$ and Spearman's $\rho$ in Table~\ref{tab:v2v_summe_tvsum}.
This suggests our steering reshapes temporal salience allocation rather than applying a global re-scaling.

\subsubsection{Mechanism from prior calibration to attention steering.}
Fig.~\ref{fig:qual_curves_ab}(B) visualizes the internal signals of \frameworkname.
Calibration converts the noisy raw highlight prior into a temporally coherent distribution $\mathbf{v}$, and the post-steering attention mass closely follows $\mathbf{v}$, indicating effective injection of the prompt-conditioned control signal into inference-time attention.
Unlike discrete extraction (score$\rightarrow$select$\rightarrow$summarize), which discards low-scored segments and creates an irreversible information bottleneck, \frameworkname retains full context and reduces conversion loss from prompt intent to evidence usage, making prompt-relevant details more accessible during generation.


\section{Conclusion}
In this work, we presented \textbf{HAS}, a highlight-guided attention steering framework for multimodal LLM video summarization. Unlike prior pipelines that rely on \emph{discrete} highlight selection, HAS turns query-conditioned highlight scores into a \emph{continuous} temporal prior and keeps the video input intact. At inference time, HAS aligns this prior to visual tokens and injects it into a small set of visual cross-attention heads to smoothly bias computation toward salient moments while avoiding the loss of low-scored but useful evidence. Experiments show consistent gains over strong baselines, improving temporal salience agreement, cross-modal consistency, and grounded long-form summarization across datasets. Looking forward, we plan to strengthen the highlight prior and make steering more adaptive, and to combine HAS with parameter-efficient adaptation to further improve long-form, faithful summarization.

\section{Societal Impact}
\label{sec:social_impact}


On the positive side, \frameworkname~can improve the accessibility of long videos by helping users navigate, retrieve, and summarize video content more efficiently.
This may be useful for educational videos, scientific presentations, and other information-dense long-form media, where grounded summarization can reduce the burden of reviewing the entire video manually.
More generally, improving evidence-aware video summarization may support better human access to large-scale video archives.

At the same time, summaries are inherently lossy.
If the highlight prior, user query, or underlying MLLM is biased or incomplete, the generated summary may over-emphasize certain moments while omitting important context.
As a result, the system could produce partial or misleading narratives, especially in high-stakes settings where factual completeness matters.
Because \frameworkname~is query-conditioned, malicious or leading prompts could also be used to generate selectively framed summaries that reinforce a desired interpretation of the video.

\section{Future Work}


First, future work can explore stronger prompt-conditioned highlight priors.
Although our calibration step smooths raw highlight scores and improves their stability, the quality of the steering signal still depends on the reliability of the initial highlight estimate.
More adaptive prior-generation strategies, uncertainty-aware calibration, or jointly learned saliency estimators may further improve robustness under out-of-domain queries, weak clip descriptions, and videos with temporally dispersed evidence.

Second, an interesting direction is to combine our coarse continuous steering prior with finer temporal localization mechanisms.
The current design intentionally avoids hard keyframe selection and instead provides stable soft guidance over the full video.
Future extensions could use a coarse-to-fine policy, where continuous attention steering preserves global context while a secondary module handles abrupt event transitions, short salient moments, or tasks requiring precise temporal boundaries.

Third, future work can further study how inference-time steering interacts with different video MLLM backbones.
Our results show that \frameworkname~can improve groundedness without updating the backbone, but the final summary quality is still influenced by the model's visual understanding ability, context budget, and instruction-following behavior.
A broader study over stronger open-source and fully fine-tuned video MLLMs may clarify when steering provides the largest benefit and how it can complement model-scale improvements.


\section*{Acknowledgments}
This research was supported in part by the National Science Foundation (NSF) SaTC-2426299 and SaTC-2413046.


\clearpage
\newpage

%
%
\bibliographystyle{splncs04}

\begingroup
  \renewcommand{\url}[1]{}
  \renewcommand{\href}[2]{#2}
  \bibliography{main}
\endgroup

\end{document}